\documentclass{article}

\PassOptionsToPackage{numbers}{natbib}

\usepackage[preprint]{neurips_2025}

\usepackage[utf8]{inputenc} 
\usepackage[T1]{fontenc}    
\usepackage{hyperref}       
\usepackage{url}            
\usepackage{booktabs}       
\usepackage{amsmath,amsfonts, extarrows}
\usepackage{nicefrac}       
\usepackage{microtype}      
\usepackage{xcolor}         
\usepackage{adjustbox}
\usepackage{multirow}
\usepackage{pifont}
\usepackage{booktabs,makecell}
\usepackage{enumitem}
\usepackage{listings}

\newcommand{\eg}{\textit{e.g.}}

\usepackage[capitalize]{cleveref}
\crefname{section}{Sec.}{Secs.}
\Crefname{section}{Section}{Sections}
\Crefname{table}{Table}{Tables}
\crefname{table}{Tab.}{Tabs.}
\crefname{figure}{Fig.}{Fig.}

\title{PanoWan: Lifting Diffusion Video Generation Models to 360$^\circ$ with Latitude/Longitude-aware Mechanisms
}

\author{%
Yifei Xia\textsuperscript{1,2,3\thanks{Equal contribution.}} \quad
Shuchen Weng\textsuperscript{4$^*$} \quad
Siqi Yang\textsuperscript{5} \quad
Jingqi Liu\textsuperscript{1,2} \quad
Chengxuan Zhu\textsuperscript{6} \\
\textbf{Minggui Teng\textsuperscript{1,2} \quad
Zijian Jia\textsuperscript{7} \quad
Han Jiang\textsuperscript{3} \quad
Boxin Shi\textsuperscript{1,2\thanks{Corresponding author.}}} \\
\textsuperscript{1}State Key Lab of Multimedia Info. Processing, School of Computer Science, Peking University \\
\textsuperscript{2}Nat'l Eng. Research Ctr. of Visual Tech., School of Computer Science, Peking University \\
\textsuperscript{3}OpenBayes Information Technology Co., Ltd. \ \ \textsuperscript{4}Beijing Academy of Artificial Intelligence \\
\textsuperscript{5}Institute for Artificial Intelligence, Peking University \\
\textsuperscript{6}Nat'l Key Lab of General AI,  School of Intelligence Science and Technology, Peking University \\
\textsuperscript{7}School of Artificial Intelligence, Beijing University of Posts and Telecommunications \\
\texttt{\{yfxia,shuchenweng,yousiki,peterzhu,minggui\_teng,shiboxin\}@pku.edu.cn} \\
\texttt{liujingqi@stu.pku.edu.cn \ \ jiazijian@bupt.edu.cn \ \ hahn@openbayes.com}
}

\begin{document}

\maketitle

\begin{abstract}
Panoramic video generation enables immersive 360° content creation, valuable in applications that demand scene-consistent world exploration.
However, existing panoramic video generation models struggle to leverage pre-trained generative priors from conventional text-to-video models for high-quality and diverse panoramic videos generation, due to limited dataset scale and the gap in spatial feature representations.
In this paper, we introduce PanoWan to effectively lift pre-trained text-to-video models to the panoramic domain, equipped with minimal modules. PanoWan employs latitude-aware sampling to avoid latitudinal distortion, while its rotated semantic denoising and padded pixel-wise decoding ensure seamless transitions at longitude boundaries.
To provide sufficient panoramic videos for learning these lifted representations, we contribute {\sc PanoVid}, a high-quality panoramic video dataset with captions and diverse scenarios.
Consequently, PanoWan achieves state-of-the-art performance in panoramic video generation and demonstrates robustness for zero-shot downstream tasks. Our project page is available at \url{https://panowan.variantconst.com}.

\end{abstract}

\section{Introduction}
\label{sec:introduction}

Text-based panoramic video generation aims to produce a complete 360° view, ensuring coherent spatial and visual relationships between elements within the scene.
Such inherent property is highly valuable for conventional VR content, the construction of interactive game worlds~\cite{xiao2025worldmem,he2025cameractrl}, and the simulation of environments for embodied AI~\cite{lu2024genex}.

The remarkable capabilities of conventional text-to-video models~\cite{veo2,wan2025,blattmann2023stable} motivate researchers to leverage their generative priors to panoramic video generation.
One intuitive strategy generates local perspective conventional videos and integrates them during inference. While these training-free methods~\cite{liu2025dynamicscaler,park2025spherediff} entirely preserve generative priors, they sacrifice overall consistency as they struggle to establish cross-view long-range dependencies.
Alternatively, fine-tuning conventional text-to-video models~\cite{wang2024360dvd,zhang2025panodit}  also faces challenges. On one hand, existing panoramic video datasets are limited in scope and scale compared to conventional ones. On the other hand, the gap in spatial representations (\eg, latitudinal distortions and seam longitudes) between panoramic and conventional videos potentially hinder the effective prior leverage from pre-trained conventional models.

\begin{figure}
  \centering
  \includegraphics[width=\linewidth]{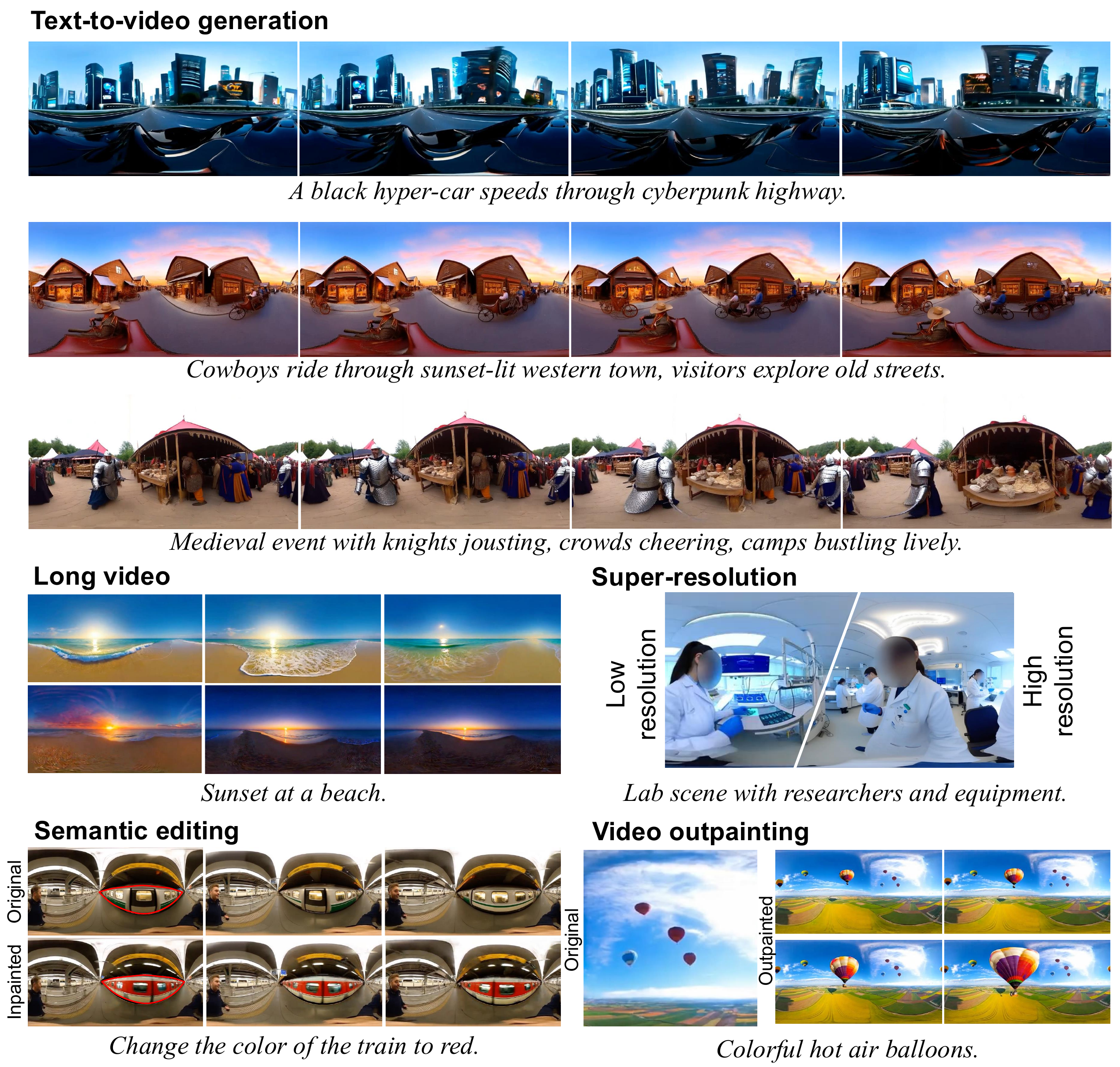}
  \caption{
  PanoWan is a text-based panoramic video generation framework. It lifts pre-trained generative priors from a conventional text-to-video model to the panorama, and enables generating diverse scenarios for long videos. 
  Equipped with training-free techniques, PanoWan supports zero-shot editing of panoramic videos, including super-resolution, semantic editing, and video outpainting.
  }
  \label{fig:teaser}
\end{figure}

In this paper, we pursue a training-based approach to overcome current bottlenecks.
We propose \textbf{PanoWan}, a framework for \textbf{Pano}ramic video generation based on \textbf{Wan} 2.1~\cite{wan2025}.
Equipped with minimal modules, PanoWan effectively lifts generative priors from a pre-trained conventional text-to-video model to the panorama.
To bridge the gap in spatial feature representations between panoramic and conventional videos, we design latitude-aware sampling to address latitudinal distortions caused by equirectangular projection. 
Since text-to-video models lack continuity awareness for left and right boundaries, we achieve seamless longitude transitions using the rotated semantic denoising to address semantic inconsistency and the padded pixel-wise decoding to resolve pixel-wise disharmony.

As learning to lift spatial representations from conventional videos to the panorama requires large-scale data, we further introduce {\sc PanoVid}. This \textbf{Pano}ramic \textbf{Vid}eo dataset offers diverse scenarios (\eg, landscape, streetscape, and humanscape), includes over 13K captioned video clips totaling 944 hours, and features data processing tailored for panoramic video generation.
As shown in~\cref{fig:teaser}, our framework produces panoramic videos from text descriptions, enabling long video generation. Additionally, it enables training-free editing of user-provided panoramic videos, including super-resolution, semantic inpainting, and video outpainting. 
Extensive experiments demonstrate that PanoWan achieves state-of-the-art panoramic video generation performance across seven metrics, alongside robust zero-shot capabilities for various downstream tasks.

Our contributions can be summarized as follows:

\begin{itemize}[align=parleft,left=0pt..1em,topsep=0pt,itemsep=0pt]
    \item We contribute {\sc PanoVid}, a large-scale and high-quality panoramic video dataset with captioned video clips, tailored for text-based panoramic video generation.
    \item We propose PanoWan, lifting generative priors from a pre-trained model to show state-of-the-art performance on panoramic video generation and robustness for downstream tasks.
    \item We integrate the latitude-aware sampling, rotated semantic denoising, and padded pixel-wise decoding to bridge the spatial difference between panoramic and conventional videos.
\end{itemize}

\section{Related Works}
\label{sec:related_works}

\subsection{Video Diffusion Models}
Diffusion models have demonstrated impressive results in image generation, but extending them to videos introduces additional challenges (\eg, temporal consistency and computational efficiency). 
Early models adapt image diffusion models with cascaded architecture~\cite{imagenvideo}, temporal layers~\cite{makeavideo}, and attention mechanisms~\cite{bertasius2021space}.
To further improve efficiency, LVDM~\cite{he2022latent} introduces a lightweight latent video diffusion model with a 3D latent space and hierarchical structure for long video generation. Meanwhile, SVD~\cite{blattmann2023stable} improves video diffusion using large and high-quality datasets. 
Recent Diffusion Transformers (DiTs) \cite{dit} effectively model complex spatio-temporal dynamics for video generation, operating on latent space compressed by 3D VAE \cite{3dvae}.
After that, more large-scale models (\eg, HunyuanVideo~\cite{hunyuanvideo}, CogVideoX~\cite{cogvideox}, and Seaweed~\cite{seawead2025seaweed}) emerge and demonstrate the benefits of scaling both model and data size.
This motivates us to adopt Wan 2.1~\cite{wan2025} as the backbone to leverage its strong generative priors and temporal modeling capabilities for panoramic video generation.

\subsection{Panoramic Video Generation}
\noindent{\textbf{Text conditioned generation.}}
360DVD~\cite{wang2024360dvd} is the pioneer to introduce stable video generation techniques to panoramic video generation by proposing a 360-Adapter and a set of 360 enhancement techniques. Training-free methods (\eg, DynamicScaler~\cite{liu2025dynamicscaler} and SphereDiff~\cite{park2025spherediff}) create panoramic videos by generating local patches and then composing them together into a complete panorama, which inherently break global consistency. 
With the development of video diffusion models, VideoPanda~\cite{xie2025videopanda} augments diffusion models with multi-view attention. PanoDiT~\cite{zhang2025panodit} uses a DiT backbone with global-temporal attention and panoramic-specific losses for coherent long-range generation. Despite these advancements, existing methods still suffer from observable latitude distortions and issues with seam-free longitude transitions. 
In this work, we introduce PanoWan, a framework that addresses these challenges by lifting generative priors from pre-trained text-to-video models to panorama.

\noindent{\textbf{Image or video conditioned generation.}} 
Imagine360~\cite{tan2024imagine360} adopts antipodal-aware motion modeling to convert perspective videos to panoramic views. Building on static panoramas, 4K4DGen~\cite{li20244k4dgen} lifts them into dynamic 4D scenes via spatial-temporal denoising. HoloTime~\cite{zhou2025holotime} further leverages Gaussian splatting and a two-stage diffusion process for high-fidelity 4D reconstruction. VidPanos~\cite{ma2024vidpanos} treats panorama generation as a space-time outpainting task from panning video inputs, while Argus~\cite{luo2025beyond} integrates motion and geometry cues for enhanced video-to-360° synthesis. 
These explorations highlight a trend toward unifying spatial, temporal, and geometric reasoning. We demonstrate that PanoWan possesses these capabilities, with robust zero-shot capabilities for downstream tasks.

\section{{\sc PanoVid} Dataset}
\label{sec:dataset}

The absence of paired datasets has long been regarded as one of the primary barriers to advancing the performance of panoramic video generation models~\cite{wang2024360dvd}.
Existing text to panoramic video generation methods~\cite{wang2024360dvd,xie2025videopanda,zhang2025panodit} mainly rely on WEB360 dataset~\cite{wang2024360dvd}, which contains only 2114 video clips of 10 seconds each. Although Argus~\cite{luo2025beyond} filters out over 283K video clips from the 360-1M dataset~\cite{wallingford2024image}, it is not built for the text-based panoramic video generation task, providing no paired captions, and showing significant distribution bias for the scenario semantics. 

To address these limitations, we present {\sc PanoVid}, a large-scale and high-quality dataset with diverse scenarios and balanced semantics, tailored for text-based panoramic video generation. 
Our data collection process begins by aggregating videos from existing panoramic sources, including 360-1M~\cite{wallingford2024image}, 360+x~\cite{chen2024x360}, Imagine360~\cite{tan2024imagine360}, WEB360~\cite{wang2024360dvd}, Panonut360~\cite{xu2024panonut}, the Miraikan 360-degree Video Dataset~\cite{miraikan360}, and a public dataset of immersive VR videos~\cite{Li2017VR360DB}. 
Subsequently, we use Qwen-2.5-VL~\cite{bai2025qwen2} to generate text descriptions and predict POI (Point-of-Interest) categories for each video. 
To ensure semantic uniqueness, we perform redundancy removal based on caption similarity. Under each POI category, the videos are further filtered according to optical flow~\cite{farneback2003two} smoothness and aesthetic scores~\cite{wu2023qalign}, retaining only the top-ranked samples. 
Thanks to the filtering pipeline, {\sc PanoVid} features more than 13K high-quality video clips totaling approximately 944 hours, and is semantically diverse and balance.

\section{Method}
\label{sec:method}

Panoramic videos have a different spatial feature representation compared to conventional ones.
Inspired by GEN3C~\cite{ren2025gen3c}, we effectively preserve the generative prior of pre-trained models by equipping minimal modules and fine-tuning a small subset of parameters via LoRA~\cite{hu2022lora}.
We firstly introduce our video diffusion backbone and formulate the spherical coordinate mapping (\cref{sec:preliminaries}). Next, we propose the latitude-aware sampling to avoid latitude distortion, along with its corresponding analysis (\cref{sec:latitude}).  
Finally, we present the rotated semantic denoising and the padded pixel-wise decoding to achieve the seamless longitude transitions (\cref{sec:longitude}).

\subsection{Preliminaries} \label{sec:preliminaries}

\noindent{\textbf{Video diffusion models.}} 
We employ Wan 2.1~\cite{wan2025} as the video generation backbone, with spatial-temporal Variational AutoEncoders (VAEs) to map high-dimensional videos into compact latent codes.
The flow matching framework~\cite{lipman2022flow} is used to model a unified denoising diffusion process. Specifically, given a clear video $x$, a VAE encoder $\mathrm{E}(\cdot)$ first projects the video into the latent space $z_{1}=\mathrm{E}(x)$. 
During training, a noise $z_{0}\sim\mathcal{N}(0,I)$ is sampled, and an intermediate latent code $z_t=tz_1+(1-t)z_0$ is constructed by linearly interpolating between $z_1$ and $z_0$ at timestep $t\in[0,1]$. The training goal is to predict the ground truth velocity $v_t=\mathrm{d}z_t/\mathrm{d}t=z_1-z_0$, and the loss function is formulated as:

\begin{equation}
    \mathcal L=\mathbb E_{z_0,z_1,c_{\text{txt}},t}||u(z_t,c_{\text{txt}},t;\theta)-v_t||^2,
\end{equation}

where $c_{\text{txt}}$ is the text embedding, $\theta$ is the parameters of the prediction model, and $u(z_t,c_{\text{txt}},t;\theta)$ is the predicted velocity of the model. 

\noindent{\textbf{Spherical coordinate mapping.}} 
Panorama captures a 360° view, inherently representing signals in spherical coordinates $(\varphi, \theta)$. To leverage generative priors from conventional images and videos that operate in Cartesian coordinates $(x,y)$, we employ the equirectangular projection (ERP) $\mathcal{P}_\text{ERP}$ to map between these coordinate systems for panoramic videos: 

\begin{equation}
\mathcal{P}_\text{ERP}:[2R]\times[R]\rightarrow [0, 2\pi]\times[-\frac\pi2,\frac\pi2],\quad
(x,y) \mapsto (\varphi, \theta)
     = \Bigl( 
      \frac{2x+1}{2R}\pi
     , \frac{2y+1-R}{2R}\pi
     \Bigr),
\end{equation}

where $R$ is the radius of the sphere. $\varphi$ and $\theta$ are longitude and latitude respectively.
While ERP enables the direct application of pre-trained VAEs to encode panoramic videos into latent codes for diffusion processes, it introduces extreme horizontal stretching in polar regions.
This horizontal stretching phenomenon arises from the altered representation of distances during projection, and is recognized by changes in horizontal signal frequency.
Let $\mathrm{d}s_\varphi$ and $\mathrm{d}s_\theta$ represent the infinitesimal arc lengths along lines of constant latitude and longitude, respectively. They are formulated as:

\begin{equation}
    \mathrm{d}s_\varphi = 2R\arcsin(\cos\theta \cdot \sin \frac{\mathrm{d}\varphi}{2}) =R\cos\theta \ \mathrm{d}\varphi, \quad \mathrm{d}s_\theta = R\ \mathrm{d}\theta,
\end{equation}

where $\theta$ is the latitude. 
We further consider the spherical frequency $f_\text{sph}$ (cycles per unit physical distance) and the Cartesian frequency $f_\text{car}$ (cycles per pixel in the image). 
Assuming that warping preserves content, their relationship in frequency is scaled by the change in distance:

\begin{equation}
    f_{\text{car}, y}(x) = f_{\text{sph}, \theta}(\varphi) \frac{\mathrm{d}s_\theta}{\mathrm{d}\theta} = Rf_{\text{sph}, \theta}(\varphi) , \quad
   f_{\text{car}, x}(y) = f_{\text{sph}, \varphi}(\theta) \frac{\mathrm{d}s_\varphi}{\mathrm{d}\varphi} = Rf_{\text{sph}, \varphi}(\theta)\cos\theta.
    \label{eq:frequency}
\end{equation}

Consequently, in polar regions ($|\theta|\approx\frac{\pi}{2}$, namely $y\approx 0$ or $y\approx R$), the horizontal frequency in the Cartesian coordinate becomes near-zero ($f_{\text{car},x}(y)\approx0$).
Such distortion in the horizontal frequency distribution significantly degrades the effectiveness of transferring priors.

\begin{figure}[t]
  \centering
  \includegraphics[width=\linewidth]{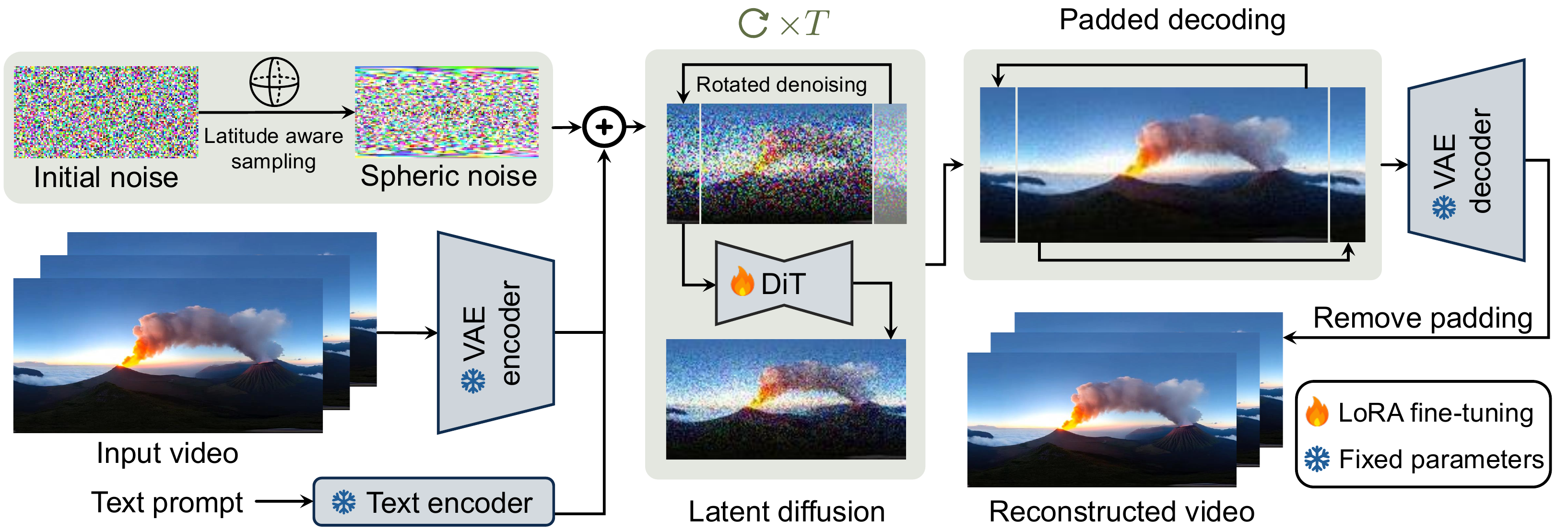}
  \caption{The pipeline of our proposed PanoWan, aware of spherical coordinates.
  To avoid latitudinal distortion, initial random Gaussian noise is remapped to align with the spherical frequency distribution using the latitude-aware sampling (\cref{sec:latitude}). 
  Next, this remapped noise serves as the latent code within the VAE-encoded latent space.
  A DiT-based denoising network then iteratively refines this latent representation, where rotated denoising is applied by rolling the latent grid to ensure semantic consistency across longitudinal boundaries.
  After that, padded pixel-wise decoding provides the VAE decoder with extended context, enabling the mapping of the denoised latent code back into seamless panoramic videos (\cref{sec:longitude}).
  The DiT backbone within PanoWan is efficiently fine-tuned using LoRA, where most parameters of the pre-trained text-to-video model remain frozen to preserve its strong generative priors.
  }

  \label{fig:method}
\end{figure}

\subsection{Latitude-Aware Mechanisms} \label{sec:latitude}

\noindent{\textbf{Latitude-aware sampling.}} 
Conventional text-to-video models typically assume independent and identically distributed (i.i.d.) Gaussian noise vectors for each Cartesian coordinate $(x,y)$. 
To avoid latitudinal distortion in polar regions of ERP, we propose the latitude-aware sampling to better align the initial noise with the spherical frequency distribution for panoramic video generation. As illustrated in the top-left of \cref{fig:method}, our latitude-aware sampling remaps the horizontal sampling coordinates based on latitude to preserve frequency consistency across the sphere. Specifically, after initializing the latent map with i.i.d. Gaussian noise vectors, we calculate the sampling noise by remapping the horizontal sampling coordinate $x$ based on the latitude corresponding to row $y$, and then applying the interpolation:

\begin{equation}
    P'(x,y) = \text{Interp}_P \Big(R+(x-R)\cos(\frac{2y+1-R}{2R}\pi),y \Big),
    \label{eq:interp}
\end{equation}

where $P'(x,y)$ is the interpolated noise vector at coordinate $(x,y)$. $\text{Interp}(\cdot)$ is formulated as the interpolation function for normalization:

\begin{equation}
    \text{Interp}_P(x,y) = \text{sgn}(\text{BI}(P,x,y)) \sqrt{\text{BI}(P^2, x,y)},
\end{equation}

where $\text{sgn}(\cdot)$ is the sign function, and $\text{BI}(P,x,y)$ is the standard bilinear interpolation for vector $P$ at coordinate $(x,y)$. 
Consequently, the resulting noise vectors sampled by our strategy preserve  $\mathbb{E} [P'(x,y)]=0$ and $\mathbb{E}[\text{Var}\ P'(x,y)]=1$, approaching the distribution on which the diffusion models are pre-trained. The proof is given in the supplementary materials.

\noindent \textbf{Frequency domain analysis.} 
We aim to prove that the horizontal frequency properties of the proposed sampling correctly represent the inherent properties of the spherical coordinate, following the methodology of 1-D Discrete Fourier Transform (DFT). 
Denote the maximal frequency of the original signal in the spherical space as $f_\text{max}$, and the maximal frequency in the Cartesian coordinates at the latitude of $\theta$ is determined by:

\begin{equation}
    \max f_{\text{car},x}(y) = \max R\cos\theta \cdot f_{\text{sph},\varphi}(\theta)\le R f_\text{max},
\end{equation}

where the equation only holds when $\theta=0$, namely on the equator. For the proposed design, it guarantees $\max f_{\text{car},x} = 2R$ along the equator as $\cos(\theta)=1$ and the original pixels along $P(\cdot, \frac{R-1}{2})$ is taken. Therefore, the maximal spherical frequency is $f_\text{max}=2$.
According to \cref{eq:interp}, the warped results only depends on the values between $(R-R\cos(\theta),y)$ and $(R+(R-1)\cos(\theta),y)$ in the original Cartesian grid $P$. According to DFT, the support of the spectrum is reduced to:

\begin{equation}
    \lceil R\cos(\theta) \rceil+\lceil (R-1)\cos(\theta) \rceil \approx 2R\cos\theta = f_\text{max}R(\theta)\cos\theta=\max f_{\text{car},x}(y), \ \forall \theta\in[-\frac{\pi}{2},\frac{\pi}{2}].
\end{equation}

This matches the inherent property of the panorama in the frequency domain in every latitude.

\subsection{Longitude Continuity Mechanisms} \label{sec:longitude}
\noindent \textbf{Seamless longitude transitions.}
Pre-trained conventional text-to-video models lack continuity awareness required between the columns of left and right boundaries. Consequently,  applying their generative priors directly for panoramic video generation leads to seam artifacts, resulting in an observable transition where the easternmost and westernmost longitudes meet. 
To achieve seamless longitude transitions, we recognize that these artifacts arise from both semantic inconsistency and pixel-wise disharmony. This motivates us to propose the rotated denoising and the padded decoding to significantly remove the artifacts.

\noindent{\textbf{Rotated semantic denoising.}} 
The video generation backbone inherently introduces semantic inconsistency at each denoising step. Since the pre-trained generative priors lack the continuity awareness, the semantic in leftmost and rightmost longitudes is typically inconsistent, which are further accumulated during the iterative denoising steps and finally produce an obvious transition. 

Our proposed rotated semantic denoising aims to spread the transition error evenly to different longitudes. 
Let $\mathcal{R}_{s_t}(\cdot)$ be the circular-shift operator and $W$ denote the width of the latent code. As shown in~\cref{fig:method}, we horizontally roll the latent code $Z_{t}$ by $\{s_t=t\bmod W\}$ columns at denoising step $t$ and then undo the shift:
\begin{equation}
Z_{t+1}
      =\mathcal{R}_{-s_t} \Bigl(\,
        \phi_\theta\bigl(\mathcal{R}_{s_t}(Z_{t})\bigr)
      \Bigr),
\end{equation}
where $\phi_\theta(\cdot)$ is the noise predictor.
As a result, the inherent accumulative error for horizontal coordinate $x$ after $T$ denoising steps is:
\begin{equation}
E_T(x)=\sum_{t=1}^{T}\varepsilon_t \bigl((x+s_t)\bmod W\bigr), 
\end{equation}
where $\varepsilon_t(\cdot)$ is prediction error for the transition and step $t$, which would concentrate at a fixed seam if no rotation are applied. 
Due to the rotation strategy, this error at physical coordinate $x$ at step $t$ is determined by the logical position $\{(x+s_t)\bmod W\}$.
Over $T$ steps, these logical coordinates $\{(x+s_t)\bmod W\}_{t=1}^T$ ideally approach a uniform permutation for all longitudes. This effectively suppresses seam artifacts by a factor approaching $1/W$.

\noindent\textbf{Padded pixel-wise decoding.} 
When decoding latent codes back to the pixel space, the pre-trained VAE decoder $\mathrm{D}$ often introduces pixel-wise inconsistencies, as it is trained on conventional videos and lacks awareness of the spatial continuity required across the left-right seam of panoramic videos~\cite{luo2025beyond}.
To address it, we present the padded pixel-wise decoding.
Let $Z_0$ be the denoised latent code. We first create a padded latent code $Z_0^\prime = P_r(Z_0)$, where $P_r(\cdot)$ is a circular padding operator that extends $Z_0$ by $r$ columns of context on the side, and the content at horizontal coordinate $x$ is $\{x \bmod W\}$ in $Z_0$. Finally, we center crop the decoded panoramic videos after the decoding $V =\operatorname{Crop} \bigl(\mathrm{D}(Z_0^\prime)\bigr)$, as illustrated in~\cref{fig:method}.
This approach ensures that pixels near the original seam boundaries are decoded with $r$ columns of horizontal panoramic context. Consequently, the VAE decoder can effectively leverage its generative priors learned from conventional videos to avoid the seam artifacts.

\section{Experiments}
\label{sec:experiments}

\subsection{Training Details}

PanoWan is built on Wan 2.1-1.3B-T2V~\cite{wan2025} as the video generation backbone. 
We train PanoWan at a resolution of $448 \times 896$, closely matching the pre-trained resolution of this backbone model. 
For parameter-efficient training, LoRA~\cite{hu2022lora} with a rank of 64 is applied to the query, key, value, and output projections of the attention mechanisms, as well as to the feed-forward networks.
The model is trained for 200K iterations on our contributed {\sc PanoVid} dataset. The training process employs the AdamW optimizer~\cite{kingma2020method} with a learning rate of $1 \times 10^{-4}$  and a batch size of 8. 
Training is conducted on 8 NVIDIA H100 GPUs for approximately 18 hours. 
During each iteration, clips of 81 consecutive frames are randomly sampled from the videos. 
Consequently, only 21.9M parameters are adjusted, constituting approximately 1.6\% of the base model's total parameters.

\subsection{Panoramic Video Evaluation Metrics}
\label{subsec:metrics}

Existing panoramic video generation methods either directly apply conventional video evaluation metrics~\cite{liu2025dynamicscaler,zhang2025panodit} or rely on subjective user preferences~\cite{wang2024360dvd}, lacking metrics that comprehensively assess both perceptual quality and spherical consistency critical for panoramic video evaluation. 
This motivates us to adapt general video quality metrics for panoramic videos and to introduce additional panorama-specific metrics for structural properties of 360° content.

\noindent{\textbf{General metrics.}} 
We apply Frechét Video Distance (FVD)~\cite{fvd} to evaluate overall video quality and VideoCLIP-XL~\cite{videoclipxl} to assess text-video alignment. Following DynamicScalar~\cite{liu2025dynamicscaler}, we also calculate specific metrics for image quality.
To adapt these general metrics for panoramic videos, we project each video onto a cube map and compute metric scores separately on each of the six faces. The final reported score for a video $v$ is a weighted average:
\begin{equation}
\overline{\mathbf{f}}(v) = \sum_{f \in \mathcal{F}} \alpha_f \cdot \Phi\big({\cal P}_f(v)\big),
\end{equation}
where $\mathcal{F}$ denotes the set of cube map faces, ${\cal P}_f(v)$ is the projection of video $v$ onto face $f \in \mathcal{F}$, $\Phi$ is the metric function, and $\alpha_f$ is the weight assigned to face $f$. Following OmniFID ~\cite{christensen2024geometry}, we assign weights $\alpha_{\text{top}} = \alpha_{\text{bottom}} = \frac{1}{3}$ and $\alpha_{\text{side}} = \frac{1}{12}$ for each of the four lateral faces.

\noindent{\textbf{Panoramic metrics.}} 
Following previous works~\cite{wang2024360dvd,liu2025dynamicscaler}, we evaluate motion patterns and scene richness with user preferences.
We additionally introduce a quantitative metric for evaluating the end continuity of generated panoramic videos, tailored to capture artifacts across longitude boundaries.
Specifically, this metric computes the mean absolute pixel difference across the left and right boundaries, directly capturing discontinuities at the longitude seam.

\begin{table*}[t]
  \centering
  \setlength\tabcolsep{10pt}
  \caption{Quantitative comparison results of PanoWan and previous text-based panoramic video generation models. $\uparrow$ ($\downarrow$) means higher (lower) is better.  Throughout the paper, best performances are highlighted in \textbf{bold}.}
  \vspace{2mm}
  \begin{adjustbox}{width=\textwidth}
  \begin{tabular}{l | ccc | ccc}
    \toprule
    \multirow[t]{2}{*}{\textbf{Method}} &
      \multicolumn{3}{c|}{\textbf{General Metrics}} &
      \multicolumn{3}{c}{\textbf{Panoramic Metrics}} \\
    \cmidrule(lr){2-4} \cmidrule(lr){5-7}
      & FVD $\downarrow$
      & \makecell{VideoCLIP- \\ XL $\uparrow$}
      & \makecell{Image \\ Quality $\uparrow$}
      & \makecell{End \\ Continuity $\downarrow$}
      & \makecell{Motion \\ Pattern $\uparrow$}
      & \makecell{Scene \\ Richness $\uparrow$} \\
    \midrule
    360DVD~\cite{wang2024360dvd}           & 1750.36 & 20.27 & 0.7054 & 0.0323 & 5.8\% & 6.6\% \\
    DynamicScaler~\cite{liu2025dynamicscaler} & 2146.04 & 21.13 & 0.7188 & 0.0339 & 4.0\% & 2.6\% \\
    \midrule
    Ours (W/o LAS) & 1520.69 & 21.20 & 0.7205 & 0.0278 & 16.2\% & 19.4\% \\
    Ours (W/o RSD) & 1302.48 & 21.76 & 0.7243 & 0.0327 & 15.6\% & 18.8\% \\
    Ours (W/o PPD) & 1294.03 & 21.81 & 0.7239 & 0.0294 & 22.0\% & 17.4\% \\
    \textbf{Ours (full)}                          & \textbf{1281.21} & \textbf{21.86} & \textbf{0.7249} & \textbf{0.0270} & \textbf{36.4\%} & \textbf{35.2\%} \\
    \bottomrule
  \end{tabular}
  \end{adjustbox}
  \label{tab:quantitative_comparison}
\end{table*}

\subsection{Comparison with State-of-the-art Methods}

\begin{figure}
  \centering
  \includegraphics[width=\linewidth]{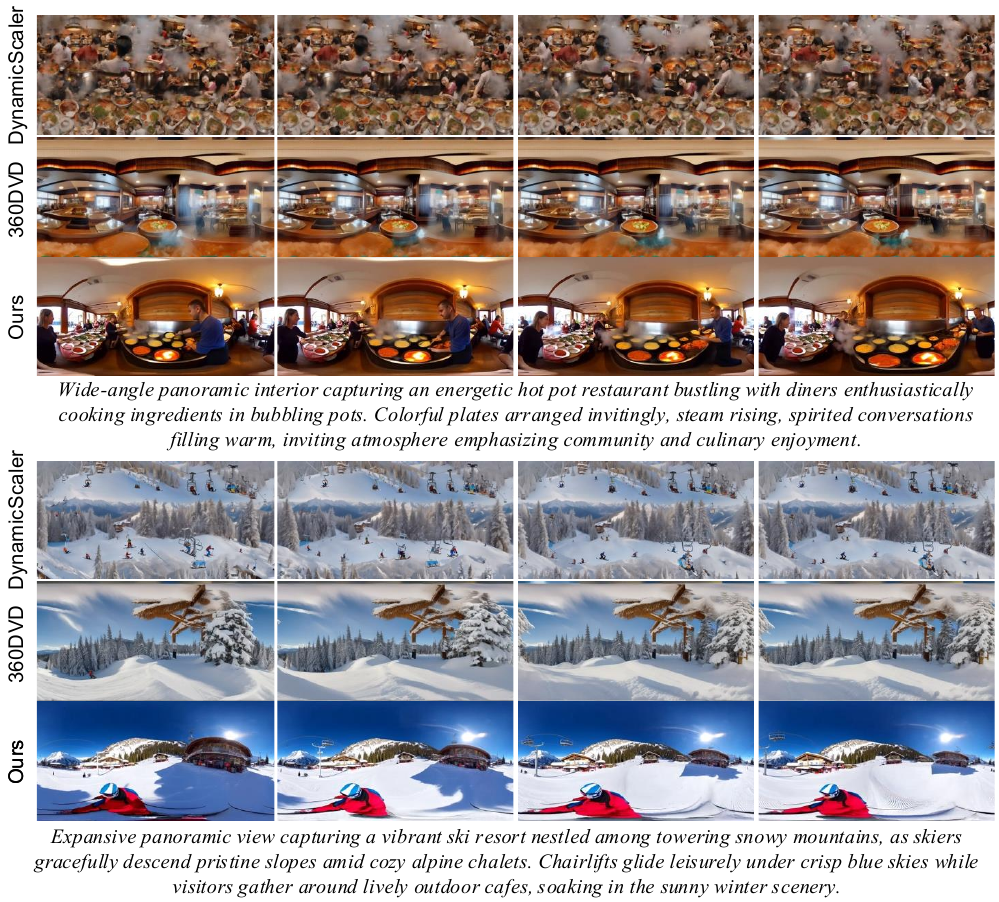}
  \caption{Visual comparison results with existing text-based panoramic video generation methods.}
  \label{fig:qualitative_comparison}
\end{figure}

\begin{figure}[t]
  \centering
  \includegraphics[width=\linewidth]{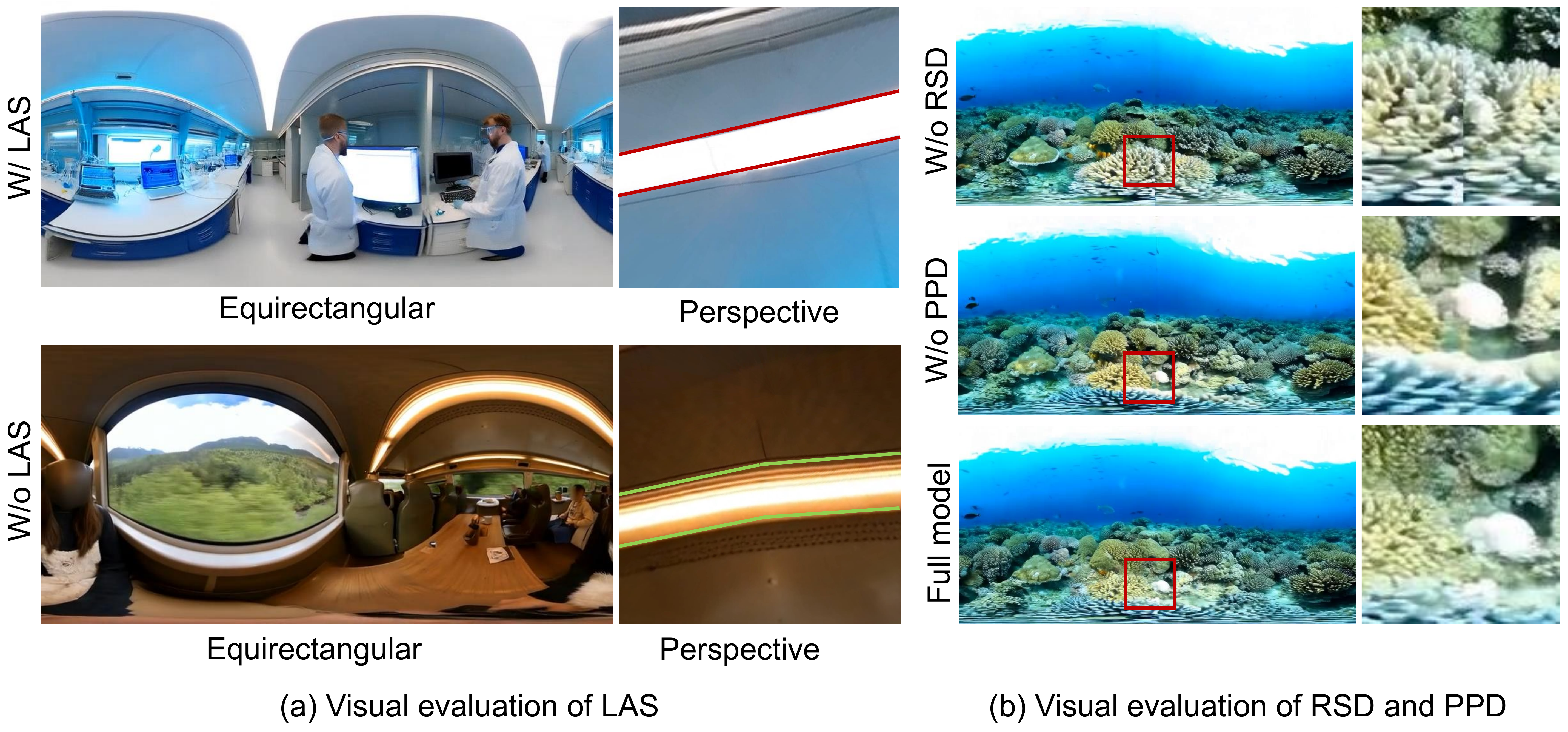}
  \caption{Qualitative evaluation of proposed latitude/longitude-aware mechanisms. 
  (a) With the proposed Latitude-Aware Sampling (LAS), PanoWan ensures that content generated at high latitudes exhibits an accurate geometry when presented in a perspective view.
  (b) By combining Rotated Semantic Denoising (RSD) and Padded Pixel-wise Decoding (PPD), PanoWan achieves seamless longitude transitions. For visualization, videos are rolled 180$^\circ$ to center the seam.}
  \label{fig:ablation}
\end{figure}

We evaluate PanoWan against existing text-based panoramic video generation methods, including 360DVD~\cite{wang2024360dvd} and DynamicScaler~\cite{liu2025dynamicscaler}. 
Quantitatively, as shown in \cref{tab:quantitative_comparison}, PanoWan achieves state-of-the-art performance across both general and panoramic metrics (detailed in \cref{subsec:metrics}).
Qualitatively, we present visual results to highlight our advantages. For instance, DynamicScaler~\cite{liu2025dynamicscaler} falls short in complex scenarios (\cref{fig:qualitative_comparison}, first sample), and 360DVD~\cite{wang2024360dvd} exhibits notable distortion in polar regions (\cref{fig:qualitative_comparison}, second sample).
In contrast, PanoWan effectively maintains global consistency and visual coherence, achieving superior performance in generating high-fidelity panoramic videos.

\subsection{Ablation Studies}
We conduct ablation studies to validate the effectiveness of proposed modules in PanoWan: latitude-aware sampling (LAS), rotated semantic denoising (RSD), and padded pixel-wise decoding (PPD). 

\noindent \textbf{Quantitative results.}
As shown in \cref{tab:quantitative_comparison}, removing LAS primarily affects general metrics—FVD increases from 1281.21 to 1520.69, and VideoCLIP-XL drops from 21.86  to 21.20—indicating the model struggles to learn panoramic features in high-latitude regions without frequency-aligned noise initialization. 
In contrast, removing RSD or PPD mainly degrades panoramic metrics (\eg, end continuity increases from 0.0270 to 0.0327  and 0.0294, respectively), confirming their roles in achieving seamless longitude transitions.

\noindent \textbf{Qualitative results.} 
We further provide qualitative evaluation in~\cref{fig:ablation}. 
When generating high-latitude elements like LED panels (which should appear straight in perspective views but are inherently distorted in the equirectangular projection), PanoWan without LAS fails to render them with the correct geometric appearance.
When RSD is discarded, semantic inconsistencies become apparent at the longitude seam, due to the lack of mechanism for continuity awareness and the error accumulation during the denoising process.
When PPD is removed, observable seam artifacts occur because conventional VAE decoder introduces pixel-wise inconsistencies at boundaries.
Consequently, the full PanoWan model with all modules enabled achieves the best performance.

\subsection{Application}
As a text-based panoramic video model, PanoWan shows robust zero-shot capabilities across a wide range of downstream tasks. 
We present representative examples in~\cref{fig:teaser} and additional examples in supplementary materials due to the space limitation.

\noindent{\textbf{Long video generation.}}  To generate long panoramic videos, we employ a local windowing strategy within the latent space. Specifically, at each denoising step, we partition the latent code into temporally overlapping chunks. These chunks are processed independently and then seamlessly merged using a linear blending function on the overlapping segments to ensure smooth temporal transitions.

\noindent{\textbf{Super-resolution for panoramic videos.}} To generate high-resolution panoramic videos from low-resolution ones, we first encode each low-resolution video into its corresponding latent code. After injected noise, the latent code is denoised based on user-provided text descriptions, producing results with structural consistency and visual fidelity across the spherical representation.

\noindent{\textbf{Inpainting for semantic editing.}}
Given a panoramic video, we identify and mask regions for modification. Next, we apply the denoising process to these regions, guided by user-provided text descriptions. Leveraging its understanding of spherical representations, the inpainted content naturally exhibits the properties of ERP projection.

\noindent{\textbf{Outpainting for conventional videos.}} 
Similar to the inpainting process, we first map the conventional video to the latent code and then mask the surrounding unseen panoramic regions. With user-provided text descriptions, we denoise the masked regions to generate corresponding content. Our pre-trained model maintains the spatial and temporal consistency for generated panoramic videos.
\section{Conclusion}
\label{sec:conclusion}

We present PanoWan, a text-based panoramic video generation framework that effectively lifts pre-trained diffusion model to the panorama. 
By integrating latitude-aware sampling, PanoWan addresses latitudinal distortions caused by equirectangular projection.
Equipped with the rotated semantic denoising and the padded pixel-wise decoding, PanoWan achieves the seamless longitude transitions.
To provide large-scale data for lifting representations from conventional videos to the panorama, we contribute the {\textsc {PanoVid}} dataset, offering high-quality and semantically rich 360° video data with annotated text descriptions.
Extensive experiments demonstrate that PanoWan achieves state-of-the-art performance on text-based panoramic video generation and strong generalization across diverse zero-shot downstream tasks.

\noindent \textbf{Limitation.}
While PanoWan benefits from the strong priors of its pretrained text-to-video models, it is also inherits a common challenge: the content forgetting problem often seen in such models. This issue is particularly evident when generating long videos due to limited temporal memory. 
We believe this challenge can be substantially alleviated through future advancements in memory-aware generation techniques (\eg, video caching mechanisms).

\setcounter{section}{6}
\setcounter{figure}{4}
\setcounter{equation}{11}
\setcounter{table}{1}
\thispagestyle{empty}

\section{Appendix}

\subsection{Proof of Noise Distribution Preservation} 

To align the initial noise with the spherical frequency distribution and avoid latitudinal distortion in polar regions of ERP, Sec.~4.2 proposes the \textit{latitude-aware sampling} strategy. 
For clarity, we recall the noise at each coordinate after remapping the horizontal sample coordinate $x$ based on latitude $y$, as stated in Eq.~(5) of the main paper:
\begin{equation}
    P'(x,y) = \text{Interp}_P \Big(R+(x-R)\cos(\frac{2y+1-R}{2R}\pi),y \Big),
    \label{eq:supp_interp}
\end{equation}

To best exploit the diffusion prior, it is desired to have $\mathbb{E} [P'(x,y)]=0$ and $\mathbb{E}[\text{Var}\ P'(x,y)]=1$. First, we provide  $\mathbb{E}[P'(x,y)]=0$ as follows:
\begin{align}
    \mathbb{E} [P'(x,y)] & = \underset{x,y}{\mathbb{E}} \left[\text{sign}(\text{BI}(P,x,y))\sqrt{\text{BI}(P^2,x,y)}\right] \\
     &=\!\! \underset{P_{ij}\sim\mathcal{N}(0,1)}{\mathbb{E}} \left[\text{sign}\left(\sum w_{ij}P_{ij} \right)\sqrt{\sum w_{ij}P_{ij}^2}\right] \\
     & \xlongequal{\tilde{P}_{ij}:=-P_{ij}}\!\!\!\!\!\!\!\!\! \underset{(-\tilde{P}_{ij})\sim\mathcal{N}(0,1)}{\mathbb{E}} \left[ \text{sign}\left(\sum w_{ij}(-\tilde{P}_{ij}) \right)\sqrt{\sum w_{ij}(-\tilde{P}_{ij})^2}\right] \\
     &=-\!\!\!\!\!\!  \underset{\tilde{P}_{ij}\sim\mathcal{N}(0,1)}{\mathbb{E}} \left[\text{sign}\left(\sum w_{ij}(P_{ij}) \right)\sqrt{\sum w_{ij}P_{ij}^2}\right] = -\mathbb{E} \left[P'(x,y)\right]. \quad\square
\end{align}
After that, we prove $\mathbb{E} [\text{Var}\ P'(x,y)]=1$ as follows:
\begin{align}
    \mathbb{E} [\text{Var}\ P'(x,y)] &= \mathbb{E}\left[P'(x,y)^2\right] - \left(\mathbb{E}\left[P'(x,y)\right]\right)^2 = \mathbb{E}\left[P'(x,y)^2\right] \\
    &= \!\! \underset{P_{ij}\sim\mathcal{N}(0,1)}{\mathbb{E}} \left[\sum_{i,j\in\{0,1\}} w_{ij}P_{ij}^2\right] = \sum_{i,j\in\{0,1\}} w_{ij}\underset{P_{ij}\sim\mathcal{N}(0,1)}{\mathbb{E}} \left[P_{ij}^2\right] \\
    &= \sum_{i,j\in\{0,1\}} \!\!w_{ij} = 1. \quad\square \label{eq:var_line2}
\end{align}
Note that the first equation in \cref{eq:var_line2} is possible only because $P_{ij}$ is independent and identically distributed.

\begin{figure}
  \centering
  \includegraphics[width=\linewidth]{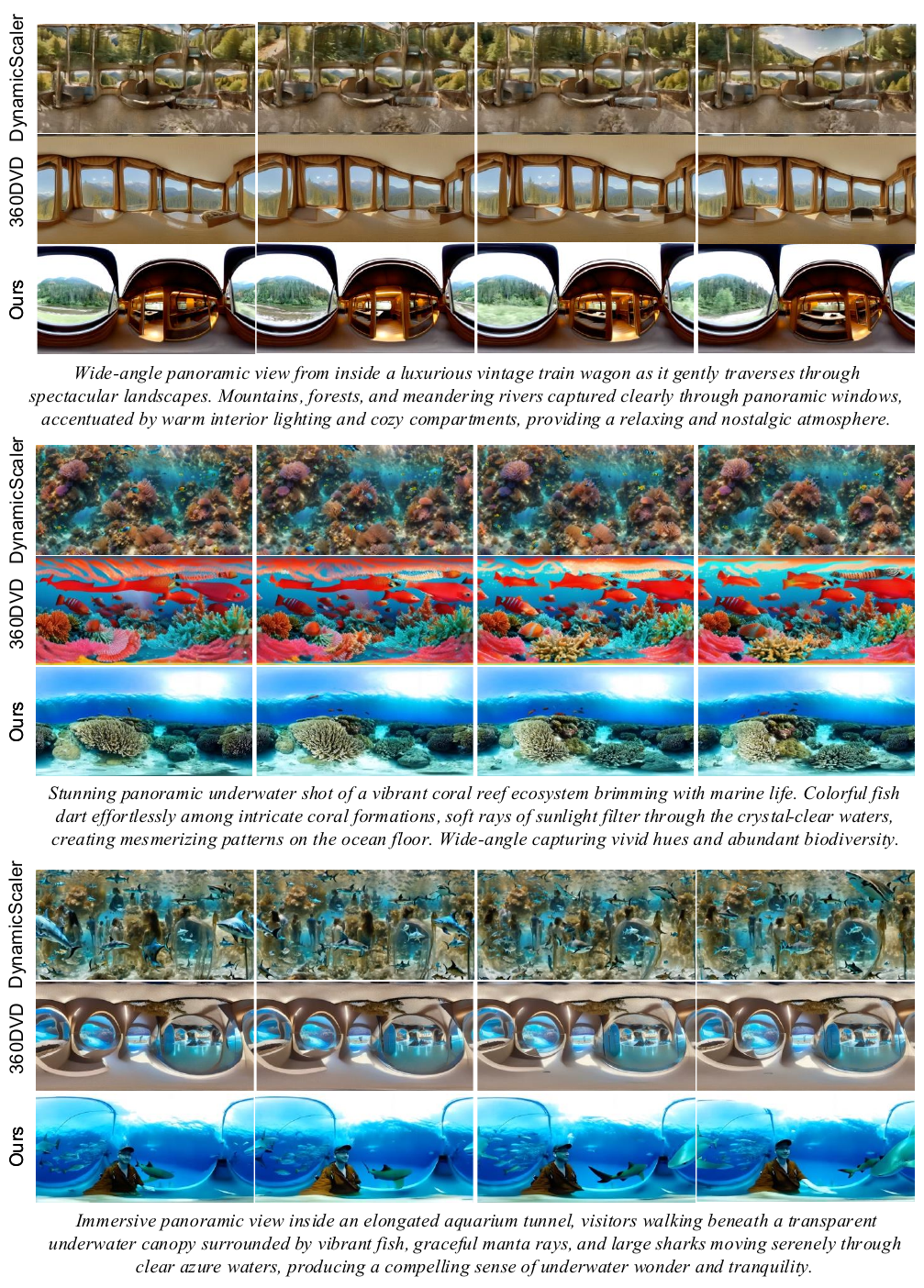}
  \caption{Additional comparison results with existing text-based panoramic video generation methods.}
  \label{fig:comparison_supp}
\end{figure}

\begin{figure}
  \centering
  \includegraphics[width=\linewidth]{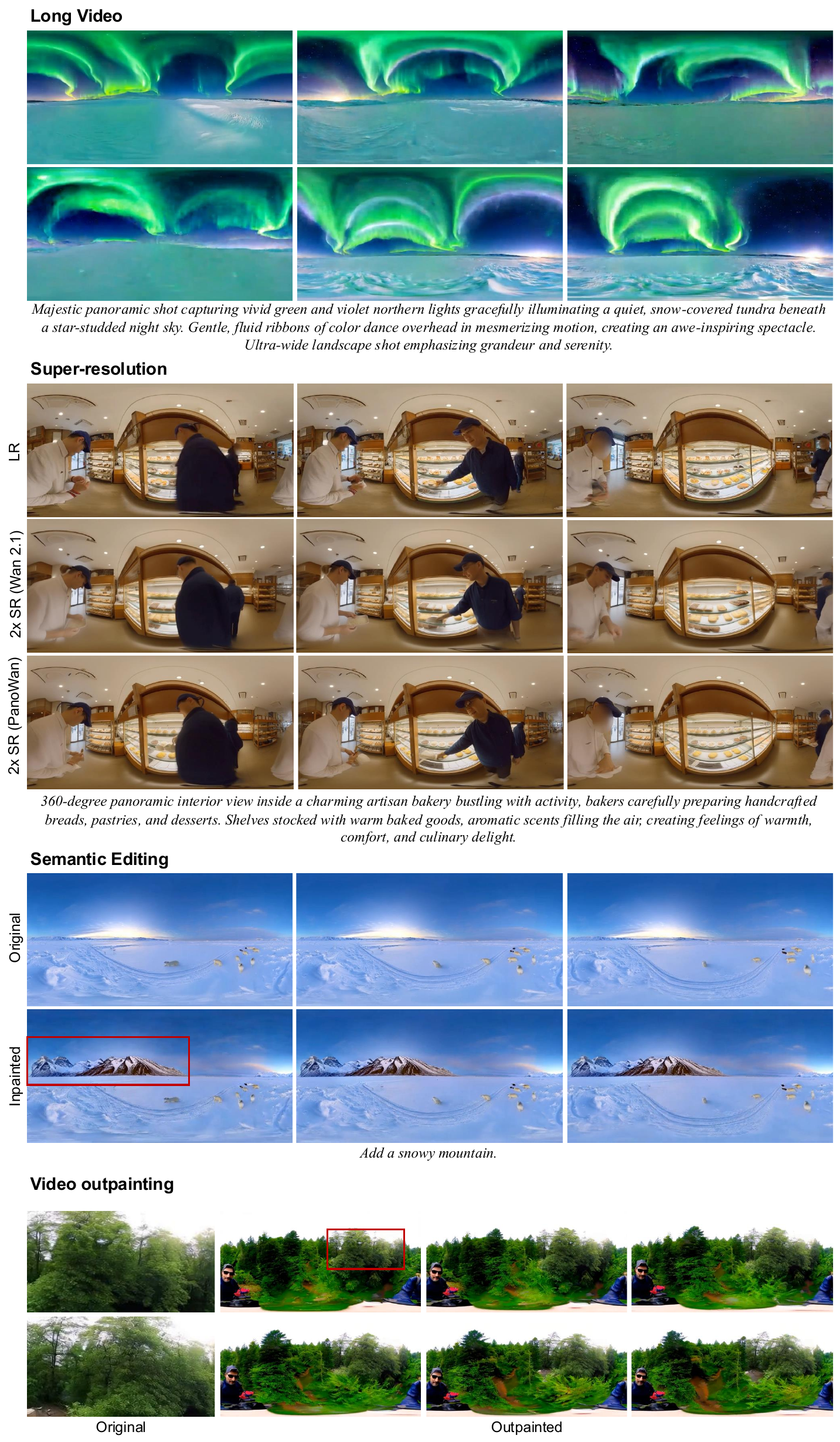}
  \caption{Additional application results, showcasing the zero-shot capabilities for downstream tasks.}
  \label{fig:application_supp}
\end{figure}

\begin{figure}
  \centering
  \includegraphics[width=\linewidth]{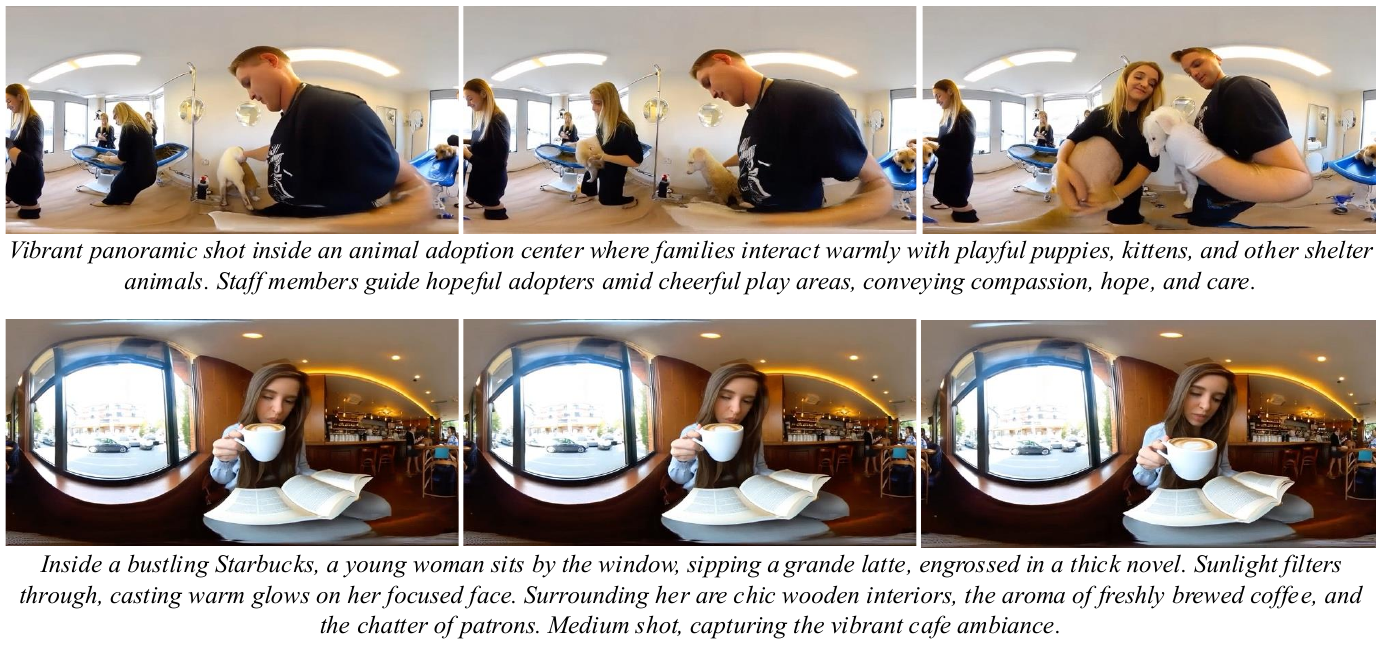}
  \caption{Visualization of failure cases.}
  \label{fig:failure_supp}
\end{figure}

\subsection{Additional Experiment Results}
In this section, we provide additional comparison results between PanoWan and existing text-based panoramic video generation methods~\cite{liu2025dynamicscaler,wang2024360dvd}, where PanoDiT~\cite{zhang2025panodit} is omitted as its code is unavailable. 
We also present additional examples showcasing applications such as long video generation, super-resolution, semantic inpainting, and video outpainting. 
Finally, we provide a detailed discussion of failure cases.

\noindent{\textbf{Additional comparison results.}} 
As shown in \cref{fig:comparison_supp}, DynamicScaler~\cite{liu2025dynamicscaler} suffers from a limited local denoising window, resulting in globally inconsistent content with repeated semantic elements appearing in different regions. 
360DVD~\cite{wang2024360dvd} often produces observable artifacts in high-latitude regions and exhibits relatively limited scene consistency. 
In contrast, our PanoWan achieves the most coherent and visually consistent results across diverse scenarios.

\noindent{\textbf{Additional application results.}}
We provide additional examples across four application scenarios.
As shown in \cref{fig:application_supp}, (a) PanoWan enables long video generation while maintaining consistent semantics throughout extended temporal durations. 
(b) For the super-resolution, directly applying Wan 2.1~\cite{wan2025} leads to severe artifacts in high-latitude regions, whereas PanoWan produces consistent and artifact-free results across all latitudes. 
(c) and (d) further demonstrate its effectiveness in semantic editing and video outpainting, respectively. 
These results highlight the strong potential of PanoWan as a versatile model for high-quality panoramic video generation and editing.

\noindent{\textbf{Failure cases.}}
PanoWan can occasionally exhibit failures in certain scenarios. 
As illustrated in the top row of~\cref{fig:failure_supp}, the generated pet dog exhibits inconsistent features. 
In the bottom row, the book being read by the woman appears with two spines and three pages, deviating from real-world structure and common sense.
We believe these failure cases are not primarily related to panoramic properties but are largely inherited from the pre-trained text-to-video model~\cite{wan2025}. As the backbone model improves, these failure cases will be improved.

\subsection{Dataset Details}

Existing panoramic video datasets~\cite{wang2024360dvd,tan2024imagine360,luo2025beyond} are limited in scale or lack paired text captions.
Recognizing the need for a dataset that supports effective and scalable training of text-based panoramic generation models, we collected our {\sc PanoVid} dataset.
In this section, we first present the data sources, followed by an efficient and accurate filtering pipeline designed to support the scalable collection of panoramic videos with diverse and paired captions.
After that, we provide an analysis of the semantic distribution and showcase additional examples from the final dataset.

\noindent \textbf{Data collection.}
We incorporate videos from large-scale collections (\eg, 360-1M~\cite{wallingford2024image} and Imagine360~\cite{tan2024imagine360}), which are primarily composed of user-uploaded YouTube content. These sources offer vast quantity of data but relatively lower visual quality.
To compensate for these quality concerns, more curated datasets are incorporated (\eg, WEB360~\cite{wang2024360dvd}, 360+x~\cite{chen2024x360}, Panonut360~\cite{xu2024panonut}, and the Miraikan 360-degree Video Dataset~\cite{miraikan360}), which are fine-grained but limited in volume.

\begin{figure}
  \centering
  \includegraphics[width=\linewidth]{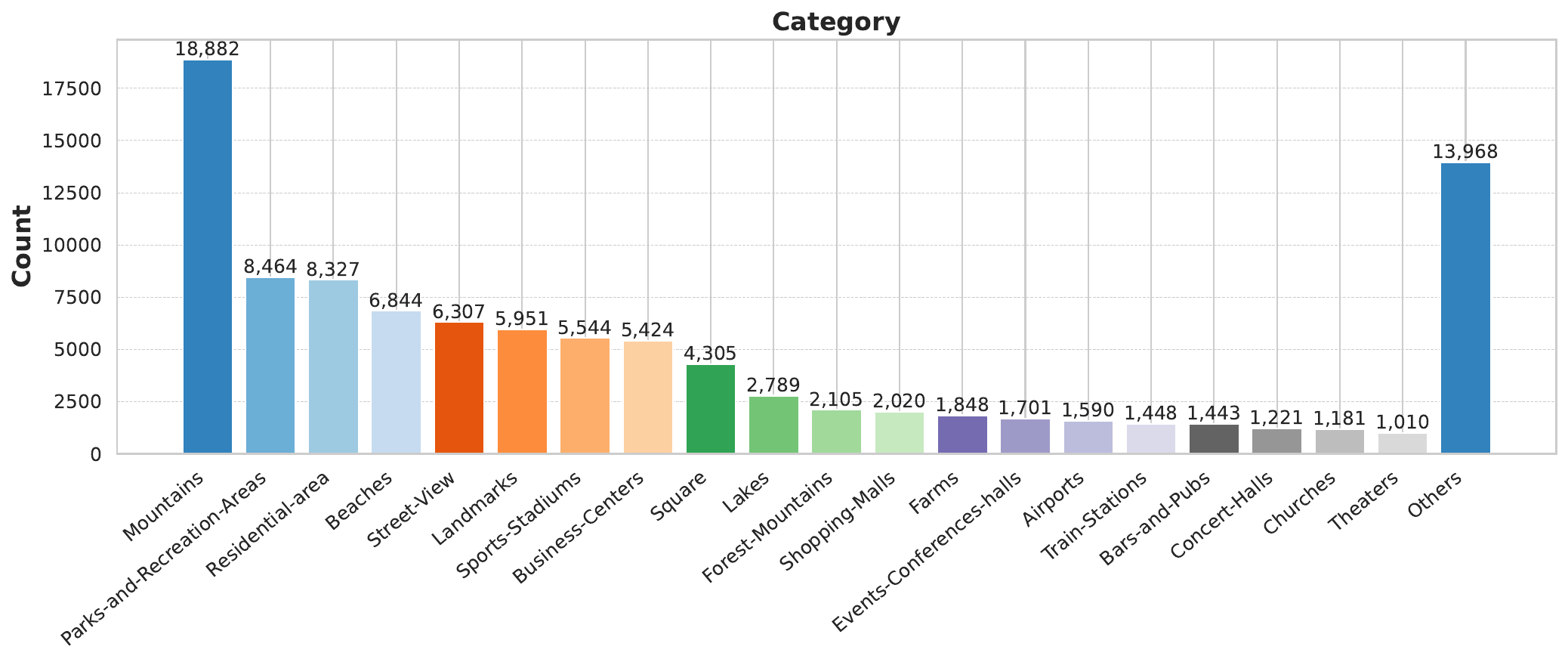}
  \caption{Category distribution of \textsc{PanoVid} dataset before balancing the semantics.}
  \label{fig:category_distribution_supp}
\end{figure}

\noindent \textbf{Data filtering.}
To efficiently filter and curate high-quality panoramic video clips from these varied noisy sources, we design a five-stage filtering pipeline based on a vision-language model. This pipeline ensures that only relevant and high-quality content is selected for further processing:

\begin{itemize}[align=parleft,left=0pt..1em,topsep=0pt,itemsep=0pt]
    \item \noindent{\textbf{Initial filtering by popularity.}} From the 360-1M dataset, we retain only videos with at least 1000 views. This heuristic filters out low-engagement content, which is often associated with poor quality or uninformative scenes.

    \item \noindent{\textbf{Video segmentation.}} We segment each video into 10-second clips using the PySceneDetect library, ensuring that each clip contains a single continuous scene without abrupt transitions.

    \item \noindent{\textbf{Vision-language-based annotation.}} For each segmented clip, we employ the Qwen-2.5-VL~\cite{bai2025qwen2} model to generate structured annotations in JSON format. The model receives the following prompt:

    \begin{lstlisting}
        Please analyze this video and provide your response in JSON format with the following fields:

        1. `caption': A detailed description of what's happening in the video. Do not show your analysis, just describe what you see. Do not start with "The video shows", describe the video itself as a whole. Include the panoramic statement like "Panoramic view of ..." or "360 degree view of ..." if it is a panoramic video. Here are some examples:
           - `Panoramic shot of colorful hot air balloons gracefully ascend, floating over lush green fields, their vibrant hues contrasting against a vast, cloud-dappled blue sky. Gentle breezes propel them in a serene dance, casting dynamic shadows on the verdant landscape below. Wide shot from ground level, capturing the expansive scene.'
           - `Panoramic shot of an active volcano spewing smoky plumes against a fiery sunset sky, majestic mountains shrouded in misty clouds in the foreground, creating a breathtaking contrast. Camera pans slowly, capturing the vastness and awe-inspiring beauty of nature.'
           - `Aerial perspective of vibrant fireworks blossoming in the ink-black sky, casting shimmering lights over a sprawling urban landscape below. Mesmerizing pyrotechnics burst in various colors and patterns against the starless expanse, illuminating cityscapes with transient brilliance. Wide shot from a plane window, capturing the nocturnal city alive under the grand firework spectacle.'
        2. `is_panorama': A boolean (true or false) indicating whether this is a 360-degree ERP projected video.
        3. `poi_category': A list of strings indicating the points of interest in the video. If there are no points of interest, set this to an empty list. If there are multiple points of interest, describe each of them in a string, sorted by their importance. You should use the available POI categories. In case none of the provided POI categories can describe the video, you may return a succinct word in the similar pattern as given categories. For example:
           - [`Coffee-Shop']
           - [`Mountains', `Lakes']
        
        Your response should be valid JSON string, like this:
        
        ```json
        {
          "caption": "Panoramic shot of colorful hot air balloons gracefully ascend, floating over lush green fields, their vibrant hues contrasting against a vast, cloud-dappled blue sky. Gentle breezes propel them in a serene dance, casting dynamic shadows on the verdant landscape below. Wide shot from ground level, capturing the expansive scene.",
          "is_panorama": true,
          "poi_category": ["Mountains"]
        }
        '''
        
        Note that the video may be compressed with limited fps to reduce uploaded file size. Your response in `caption' should not include any description of the video quality or compression. Just focus on the content of the video.
        
        Here are the available POI categories: "Restaurant, Coffee-Shop, Bars-and-Pubs, Residential-area, Hotels-Motels, Vaccation-Rentals, Hospitals-Clinics, Pharmacies, Dentists, School-Universities, Library, Supermarkets, Shopping-Malls, Clothing-Stores, Shoe-Stores, Bookstores, Flowerstore, Furniture-Stores, Electorical-Store, Pet-Store, Toy-Shop, Airports, Train-Stations, Bus-Stops, Gas-Station, Car-Rental-Agencies, Theaters, Concert-Halls, Sports-Stadiums, Parks-and-Recreation-Areas, Museums, Art-Galleries, Zoos-Aquariums, Botanical-Gardens, Landmarks, Cultural-Centers, Post-Offices, Police-Stations, Courthouses, City-Halls, Banks-ATMs, Events-Conferences-halls, Beaches, Hiking-Trails, Campgrounds, Lakes, Mountains, Forest-Mountains, Farms, Street-View, Square, Business-Centers, Tech-Companies, Co-working-Spaces, Gyms-and-Fitness-Centers, Sports-Clubs, Swimming-Pools, Tennis-Courts, Auto-Repair-Shops, Car-Washes, Parking-Lots, Churches, Mosques, Temples, Graveyards."
    \end{lstlisting}

    The model’s responses are used to filter out clips that are not panoramic videos in ERP format. Note that the POI (Points Of Interest) categories follow the setup of DL3DV~\cite{ling2024dl3dv}.

    \item \noindent{\textbf{Motion score filtering.}} We compute a normalized motion score for each clip based on optical flow magnitude (following~\cite{farneback2003two}). Only clips with motion scores above 0.4 are retained, to avoid including static or nearly still scenes that are less informative for training generative models.

    \item \noindent{\textbf{Aesthetic score filtering.}} Each frame is scored using Q-Align~\cite{wu2023qalign} for visual aesthetics, and the clip's final score is computed as the average of all frame scores. We discard clips with aesthetic scores below 3, ensuring the training data maintains a minimum level of visual quality.
\end{itemize}

This scalable filtering pipeline enables the construction of a large, diverse, and high-quality panoramic video dataset from noisy web-scale sources, thereby effectively lifting pre-trained text-to-video generation models to the panorama.

\noindent{\textbf{Semantic distribution.}}
We observe that the POI labels generated through the proposed pipeline exhibit a highly imbalanced distribution, as shown in \cref{fig:category_distribution_supp}. A few categories (\eg, \textit{Mountains}, \textit{Parks-and-Recreation-Areas}, and \textit{Residential-area}) dominate the dataset with thousands of samples, while many others (\eg, \textit{Bookstores}, \textit{Dentists}, and \textit{Tennis-Courts}) have fewer than 50 instances. This imbalanced distribution poses a challenge for training panoramic video generation models that require semantic diversity and balanced representation. 
To address this issue and ensure a more balanced semantic coverage, we select up to 200 video clips with the highest aesthetic quality for each category. For categories with fewer than 200 clips, all available samples are retained. By combining this filtered set with existing curated datasets, \textsc{PanoVid} contains over 13K high-quality clips with diverse scene types and detailed captions.

\begin{figure}
  \centering
  \includegraphics[width=\linewidth]{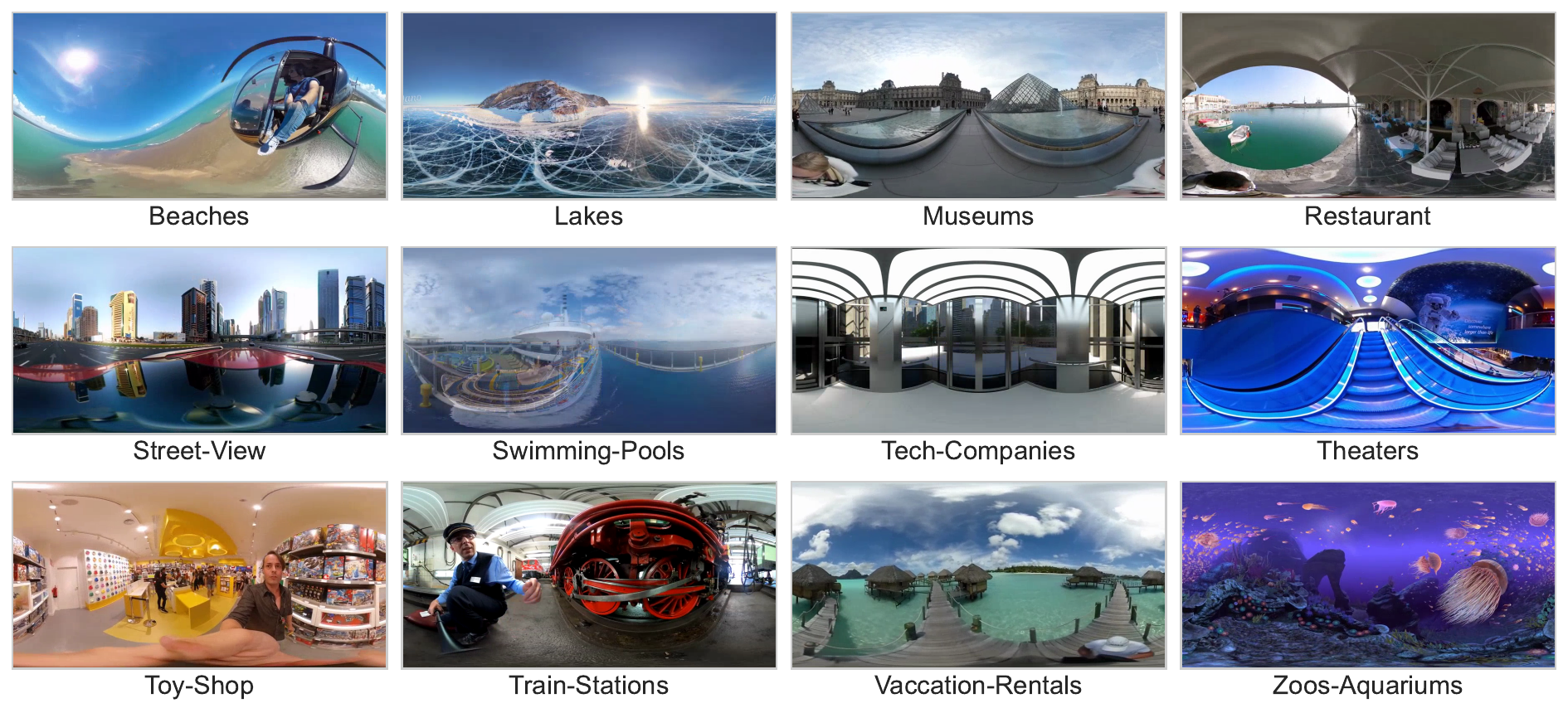}
  \caption{Representative samples from the \textsc{PanoVid} dataset.}
  \label{fig:dataset_examples_supp}
\end{figure}

\noindent{\textbf{Dataset examples.}}
\textsc{PanoVid} dataset covers a wide range of scene categories (\eg, natural landscapes, urban environments, and indoor scenarios), with paired detailed captions. 
As shown in~\cref{fig:dataset_examples_supp}, we provide representative samples from the \textsc{PanoVid} dataset to illustrate its diversity and quality.

\bibliographystyle{plainnat}
\bibliography{neurips_2025}

\begin{thebibliography}{41}
\providecommand{\natexlab}[1]{#1}
\providecommand{\url}[1]{\texttt{#1}}
\expandafter\ifx\csname urlstyle\endcsname\relax
  \providecommand{\doi}[1]{doi: #1}\else
  \providecommand{\doi}{doi: \begingroup \urlstyle{rm}\Url}\fi

\bibitem[mir()]{miraikan360}
Miraikan 360-degree video dataset.
\newblock \url{https://www.miraikan.jst.go.jp/en/research/AccessibilityLab/dataset360/}.

\bibitem[Bai et~al.(2025)Bai, Chen, Liu, Wang, Ge, Song, Dang, Wang, Wang, Tang, et~al.]{bai2025qwen2}
Shuai Bai, Keqin Chen, Xuejing Liu, Jialin Wang, Wenbin Ge, Sibo Song, Kai Dang, Peng Wang, Shijie Wang, Jun Tang, et~al.
\newblock Qwen2.5-{VL} technical report.
\newblock \emph{arXiv preprint arXiv:2502.13923}, 2025.

\bibitem[Bertasius et~al.(2021)Bertasius, Wang, and Torresani]{bertasius2021space}
Gedas Bertasius, Heng Wang, and Lorenzo Torresani.
\newblock Is space-time attention all you need for video understanding?
\newblock In \emph{ICML}, 2021.

\bibitem[Blattmann et~al.(2023)Blattmann, Dockhorn, Kulal, Mendelevitch, Kilian, Lorenz, Levi, English, Voleti, Letts, et~al.]{blattmann2023stable}
Andreas Blattmann, Tim Dockhorn, Sumith Kulal, Daniel Mendelevitch, Maciej Kilian, Dominik Lorenz, Yam Levi, Zion English, Vikram Voleti, Adam Letts, et~al.
\newblock Stable video diffusion: Scaling latent video diffusion models to large datasets.
\newblock \emph{arXiv preprint arXiv:2311.15127}, 2023.

\bibitem[Chen et~al.(2024)Chen, Hou, Qu, Testini, Hong, and Jiao]{chen2024x360}
Hao Chen, Yuqi Hou, Chenyuan Qu, Irene Testini, Xiaohan Hong, and Jianbo Jiao.
\newblock 360+x: A panoptic multi-modal scene understanding dataset.
\newblock In \emph{Proc. of IEEE/CVF Conference on Computer Vision and Pattern Recognition}, 2024.

\bibitem[Christensen et~al.(2024)Christensen, Mojab, Patel, Ahuja, Akata, Winther, Gonzalez-Franco, and Colaco]{christensen2024geometry}
Anders Christensen, Nooshin Mojab, Khushman Patel, Karan Ahuja, Zeynep Akata, Ole Winther, Mar Gonzalez-Franco, and Andrea Colaco.
\newblock Geometry fidelity for spherical images.
\newblock In \emph{Proc. of European Conference on Computer Vision}, 2024.

\bibitem[Farneb{\"a}ck(2003)]{farneback2003two}
Gunnar Farneb{\"a}ck.
\newblock Two-frame motion estimation based on polynomial expansion.
\newblock In \emph{Image Analysis}, 2003.

\bibitem[He et~al.(2025)He, Yang, Lin, Xu, Wei, Gui, Zhao, Wetzstein, Jiang, and Li]{he2025cameractrl}
Hao He, Ceyuan Yang, Shanchuan Lin, Yinghao Xu, Meng Wei, Liangke Gui, Qi~Zhao, Gordon Wetzstein, Lu~Jiang, and Hongsheng Li.
\newblock Camera{C}trl {II}: Dynamic scene exploration via camera-controlled video diffusion models.
\newblock \emph{arXiv preprint arXiv:2503.10592}, 2025.

\bibitem[He et~al.(2022)He, Yang, Zhang, Shan, and Chen]{he2022latent}
Yingqing He, Tianyu Yang, Yong Zhang, Ying Shan, and Qifeng Chen.
\newblock Latent video diffusion models for high-fidelity long video generation.
\newblock \emph{arXiv preprint arXiv:2211.13221}, 2022.

\bibitem[Ho et~al.(2022)Ho, Chan, Saharia, Whang, Gao, Gritsenko, Kingma, Poole, Norouzi, Fleet, and Salimans]{imagenvideo}
Jonathan Ho, William Chan, Chitwan Saharia, Jay Whang, Ruiqi Gao, Alexey Gritsenko, Diederik~P Kingma, Ben Poole, Mohammad Norouzi, David~J Fleet, and Tim Salimans.
\newblock Imagen video: High definition video generation with diffusion models.
\newblock \emph{arXiv preprint arXiv:2210.02303}, 2022.

\bibitem[Hu et~al.(2022)Hu, Shen, Wallis, Allen-Zhu, Li, Wang, Wang, Chen, et~al.]{hu2022lora}
Edward~J Hu, Yelong Shen, Phillip Wallis, Zeyuan Allen-Zhu, Yuanzhi Li, Shean Wang, Lu~Wang, Weizhu Chen, et~al.
\newblock Lora: Low-rank adaptation of large language models.
\newblock In \emph{Proc. of the International Conference on Learning Representations}, 2022.

\bibitem[Jinxiu et~al.(2025)Jinxiu, Shaoheng, Yinxiao, and Ming-Hsuan]{liu2025dynamicscaler}
Liu Jinxiu, Lin Shaoheng, Li~Yinxiao, and Yang Ming-Hsuan.
\newblock Dynamic{S}caler: Seamless and scalable video generation for panoramic scenes.
\newblock In \emph{Proc. of IEEE/CVF Conference on Computer Vision and Pattern Recognition}, 2025.

\bibitem[Kingma et~al.(2020)Kingma, Ba, and Adam]{kingma2020method}
Diederik~P Kingma, J~Adam Ba, and J~Adam.
\newblock A method for stochastic optimization.
\newblock \emph{arXiv preprint arXiv:1412.6980}, 2020.

\bibitem[Kong et~al.(2024)Kong, Tian, Zhang, Min, Dai, Zhou, Xiong, Li, Wu, Zhang, et~al.]{hunyuanvideo}
Weijie Kong, Qi~Tian, Zijian Zhang, Rox Min, Zuozhuo Dai, Jin Zhou, Jiangfeng Xiong, Xin Li, Bo~Wu, Jianwei Zhang, et~al.
\newblock Hunyuanvideo: A systematic framework for large video generative models.
\newblock \emph{arXiv preprint arXiv:2412.03603}, 2024.

\bibitem[Li et~al.(2017)Li, Bailenson, Pines, Greenleaf, and Williams]{Li2017VR360DB}
Benjamin~J. Li, Jeremy~N. Bailenson, Adam Pines, Walter~J. Greenleaf, and Leanne~M. Williams.
\newblock A public database of immersive vr videos with corresponding ratings of arousal, valence, and correlations between head movements and self report measures.
\newblock \emph{Frontiers in Psychology}, 2017.

\bibitem[Li et~al.(2024)Li, Pan, Yang, Xu, Zhou, Zhang, Li, Kadambi, Wang, Tu, et~al.]{li20244k4dgen}
Renjie Li, Panwang Pan, Bangbang Yang, Dejia Xu, Shijie Zhou, Xuanyang Zhang, Zeming Li, Achuta Kadambi, Zhangyang Wang, Zhengzhong Tu, et~al.
\newblock 4{K}4{DG}en: Panoramic 4{D} generation at 4{K} resolution.
\newblock \emph{arXiv preprint arXiv:2406.13527}, 2024.

\bibitem[Ling et~al.(2024)Ling, Sheng, Tu, Zhao, Xin, Wan, Yu, Guo, Yu, Lu, et~al.]{ling2024dl3dv}
Lu~Ling, Yichen Sheng, Zhi Tu, Wentian Zhao, Cheng Xin, Kun Wan, Lantao Yu, Qianyu Guo, Zixun Yu, Yawen Lu, et~al.
\newblock {DL3DV}-10{K}: A large-scale scene dataset for deep learning-based 3{D} vision.
\newblock In \emph{Proceedings of the IEEE/CVF Conference on Computer Vision and Pattern Recognition}, pages 22160--22169, 2024.

\bibitem[Lipman et~al.(2022)Lipman, Chen, Ben-Hamu, Nickel, and Le]{lipman2022flow}
Yaron Lipman, Ricky~TQ Chen, Heli Ben-Hamu, Maximilian Nickel, and Matt Le.
\newblock Flow matching for generative modeling.
\newblock \emph{arXiv preprint arXiv:2210.02747}, 2022.

\bibitem[Lu et~al.(2025)Lu, Shu, Xiao, Ye, Wang, Peng, Wei, Khashabi, Chellappa, Yuille, and Chen]{lu2024genex}
Taiming Lu, Tianmin Shu, Junfei Xiao, Luoxin Ye, Jiahao Wang, Cheng Peng, Chen Wei, Daniel Khashabi, Rama Chellappa, Alan Yuille, and Jieneng Chen.
\newblock Gen{E}x: Generating an explorable world.
\newblock \emph{Proc. of the International Conference on Learning Representations}, 2025.

\bibitem[Luo et~al.(2025)Luo, Wallingford, Farhadi, Snavely, and Ma]{luo2025beyond}
Rundong Luo, Matthew Wallingford, Ali Farhadi, Noah Snavely, and Wei-Chiu Ma.
\newblock Beyond the frame: Generating 360 $\!^\circ$ panoramic videos from perspective videos.
\newblock \emph{arXiv preprint arXiv:2504.07940}, 2025.

\bibitem[Ma et~al.(2024)Ma, Lu, Paiss, Zada, Holynski, Dekel, Curless, Rubinstein, and Cole]{ma2024vidpanos}
Jingwei Ma, Erika Lu, Roni Paiss, Shiran Zada, Aleksander Holynski, Tali Dekel, Brian Curless, Michael Rubinstein, and Forrester Cole.
\newblock Vid{P}anos: Generative panoramic videos from casual panning videos.
\newblock In \emph{Proc. of ACM SIGGRAPH Asia}, 2024.

\bibitem[Park et~al.(2025)Park, Kang, Yun, Hwang, and Choo]{park2025spherediff}
Minho Park, Taewoong Kang, Jooyeol Yun, Sungwon Hwang, and Jaegul Choo.
\newblock Sphere{D}iff: Tuning-free omnidirectional panoramic image and video generation via spherical latent representation.
\newblock \emph{arXiv preprint arXiv:2504.14396}, 2025.

\bibitem[Peebles and Xie(2023)]{dit}
William Peebles and Saining Xie.
\newblock Scalable diffusion models with transformers.
\newblock In \emph{ICCV}, 2023.

\bibitem[Ren et~al.(2025)Ren, Shen, Huang, Ling, Lu, Nimier-David, M{\"u}ller, Keller, Fidler, and Gao]{ren2025gen3c}
Xuanchi Ren, Tianchang Shen, Jiahui Huang, Huan Ling, Yifan Lu, Merlin Nimier-David, Thomas M{\"u}ller, Alexander Keller, Sanja Fidler, and Jun Gao.
\newblock {GEN3C}: 3{D}-informed world-consistent video generation with precise camera control.
\newblock In \emph{Proc. of IEEE/CVF Conference on Computer Vision and Pattern Recognition}, 2025.

\bibitem[SeaweadTeam et~al.(2025)SeaweadTeam, Yang, Lin, Zhao, Lin, Ma, Guo, Chen, Qi, Wang, et~al.]{seawead2025seaweed}
SeaweadTeam, Ceyuan Yang, Zhijie Lin, Yang Zhao, Shanchuan Lin, Zhibei Ma, Haoyuan Guo, Hao Chen, Lu~Qi, Sen Wang, et~al.
\newblock Seaweed-7{B}: Cost-effective training of video generation foundation model.
\newblock \emph{arXiv preprint arXiv:2504.08685}, 2025.

\bibitem[Singer et~al.(2023)Singer, Polyak, Hayes, Yin, An, Zhang, Hu, Yang, Ashual, Gafni, et~al.]{makeavideo}
Uriel Singer, Adam Polyak, Thomas Hayes, Xi~Yin, Jie An, Songyang Zhang, Qiyuan Hu, Harry Yang, Oron Ashual, Oran Gafni, et~al.
\newblock Make-a-video: Text-to-video generation without text-video data.
\newblock In \emph{ICLR}, 2023.

\bibitem[Tan et~al.(2024)Tan, Yang, Wu, He, Guo, Liu, and Lin]{tan2024imagine360}
Jing Tan, Shuai Yang, Tong Wu, Jingwen He, Yuwei Guo, Ziwei Liu, and Dahua Lin.
\newblock Imagine360: Immersive 360 video generation from perspective anchor.
\newblock \emph{arXiv preprint arXiv:2412.03552}, 2024.

\bibitem[Unterthiner et~al.(2018)Unterthiner, Van~Steenkiste, Kurach, Marinier, Michalski, and Gelly]{fvd}
Thomas Unterthiner, Sjoerd Van~Steenkiste, Karol Kurach, Raphael Marinier, Marcin Michalski, and Sylvain Gelly.
\newblock Towards accurate generative models of video: A new metric \& challenges.
\newblock \emph{arXiv preprint arXiv:1812.01717}, 2018.

\bibitem[Veo-Team et~al.(2024)Veo-Team, :, Gupta, Razavi, Toor, Gupta, Erhan, Shaw, Lau, Belletti, Barth-Maron, Shaw, Erdogan, Sidahmed, Nandwani, Moraldo, Kim, Blok, Donahue, Lezama, Mathewson, David, Lorrain, van Zee, Narasimhan, Wang, Babaeizadeh, Papalampidi, Pezzotti, Jha, Barnes, Kindermans, Hornung, Villegas, Poplin, Zaiem, Dieleman, Ebrahimi, Wisdom, Zhang, Fruchter, Nørly, Hua, Yan, Du, and Chen]{veo2}
Veo-Team, :, Agrim Gupta, Ali Razavi, Andeep Toor, Ankush Gupta, Dumitru Erhan, Eleni Shaw, Eric Lau, Frank Belletti, Gabe Barth-Maron, Gregory Shaw, Hakan Erdogan, Hakim Sidahmed, Henna Nandwani, Hernan Moraldo, Hyunjik Kim, Irina Blok, Jeff Donahue, José Lezama, Kory Mathewson, Kurtis David, Matthieu~Kim Lorrain, Marc van Zee, Medhini Narasimhan, Miaosen Wang, Mohammad Babaeizadeh, Nelly Papalampidi, Nick Pezzotti, Nilpa Jha, Parker Barnes, Pieter-Jan Kindermans, Rachel Hornung, Ruben Villegas, Ryan Poplin, Salah Zaiem, Sander Dieleman, Sayna Ebrahimi, Scott Wisdom, Serena Zhang, Shlomi Fruchter, Signe Nørly, Weizhe Hua, Xinchen Yan, Yuqing Du, and Yutian Chen.
\newblock Veo 2.
\newblock 2024.
\newblock URL \url{https://deepmind.google/technologies/veo/veo-2/}.

\bibitem[Wallingford et~al.(2024)Wallingford, Bhattad, Kusupati, Ramanujan, Deitke, Kembhavi, Mottaghi, Ma, and Farhadi]{wallingford2024image}
Matthew Wallingford, Anand Bhattad, Aditya Kusupati, Vivek Ramanujan, Matt Deitke, Aniruddha Kembhavi, Roozbeh Mottaghi, Wei-Chiu Ma, and Ali Farhadi.
\newblock From an image to a scene: Learning to imagine the world from a million 360° videos.
\newblock In \emph{Proc. of Neural Information Processing Systems}, 2024.

\bibitem[Wang et~al.(2024{\natexlab{a}})Wang, Wang, Huang, Huang, and Jin]{videoclipxl}
Jiapeng Wang, Chengyu Wang, Kunzhe Huang, Jun Huang, and Lianwen Jin.
\newblock Video{CLIP}-{XL}: Advancing long description understanding for video clip models.
\newblock \emph{arXiv preprint arXiv:2410.00741}, 2024{\natexlab{a}}.

\bibitem[Wang et~al.(2024{\natexlab{b}})Wang, Li, Mou, Cheng, and Zhang]{wang2024360dvd}
Qian Wang, Weiqi Li, Chong Mou, Xinhua Cheng, and Jian Zhang.
\newblock 360{DVD}: Controllable panorama video generation with 360-degree video diffusion model.
\newblock In \emph{Proc. of IEEE/CVF Conference on Computer Vision and Pattern Recognition}, 2024{\natexlab{b}}.

\bibitem[WanTeam et~al.(2025)WanTeam, Wang, Ai, Wen, Mao, Xie, Chen, Yu, Zhao, Yang, et~al.]{wan2025}
WanTeam, Ang Wang, Baole Ai, Bin Wen, Chaojie Mao, Chen-Wei Xie, Di~Chen, Feiwu Yu, Haiming Zhao, Jianxiao Yang, et~al.
\newblock Wan: Open and advanced large-scale video generative models.
\newblock \emph{arXiv preprint arXiv:2503.20314}, 2025.

\bibitem[Wu et~al.(2023)Wu, Zhang, Zhang, Chen, Li, Liao, Wang, Zhang, Sun, Yan, Min, Zhai, and Lin]{wu2023qalign}
Haoning Wu, Zicheng Zhang, Weixia Zhang, Chaofeng Chen, Chunyi Li, Liang Liao, Annan Wang, Erli Zhang, Wenxiu Sun, Qiong Yan, Xiongkuo Min, Guangtai Zhai, and Weisi Lin.
\newblock Q-{A}lign: Teaching {LMM}s for visual scoring via discrete text-defined levels.
\newblock \emph{arXiv preprint arXiv:2312.17090}, 2023.

\bibitem[Xiao et~al.(2025)Xiao, Lan, Zhou, Ouyang, Yang, Zeng, and Pan]{xiao2025worldmem}
Zeqi Xiao, Yushi Lan, Yifan Zhou, Wenqi Ouyang, Shuai Yang, Yanhong Zeng, and Xingang Pan.
\newblock {WORLDMEM}: Long-term consistent world simulation with memory.
\newblock \emph{arXiv preprint arXiv:2504.12369}, 2025.

\bibitem[Xie et~al.(2025)Xie, Sabour, Huang, Paschalidou, Klar, Iqbal, Fidler, and Zeng]{xie2025videopanda}
Kevin Xie, Amirmojtaba Sabour, Jiahui Huang, Despoina Paschalidou, Greg Klar, Umar Iqbal, Sanja Fidler, and Xiaohui Zeng.
\newblock Video{P}anda: Video panoramic diffusion with multi-view attention.
\newblock \emph{arXiv preprint arXiv:2504.11389}, 2025.

\bibitem[Xu et~al.(2024)Xu, Du, Wang, Ning, Zhou, and Cao]{xu2024panonut}
Yutong Xu, Junhao Du, Jiahe Wang, Yuwei Ning, Sihan Zhou, and Yang Cao.
\newblock Panonut360: A head and eye tracking dataset for panoramic video.
\newblock In \emph{Proceedings of the 15th ACM Multimedia Systems Conference}, 2024.

\bibitem[Yang et~al.(2025)Yang, Teng, Zheng, Ding, Huang, Xu, Yang, Hong, Zhang, Feng, et~al.]{cogvideox}
Zhuoyi Yang, Jiayan Teng, Wendi Zheng, Ming Ding, Shiyu Huang, Jiazheng Xu, Yuanming Yang, Wenyi Hong, Xiaohan Zhang, Guanyu Feng, et~al.
\newblock Cog{V}ideox: Text-to-video diffusion models with an expert transformer.
\newblock In \emph{ICLR}, 2025.

\bibitem[Yu et~al.(2024)Yu, Lezama, Gundavarapu, Versari, Sohn, Minnen, Cheng, Birodkar, Gupta, Gu, Hauptmann, Gong, Yang, Essa, Ross, and Jiang]{3dvae}
Lijun Yu, José Lezama, Nitesh~B. Gundavarapu, Luca Versari, Kihyuk Sohn, David Minnen, Yong Cheng, Vighnesh Birodkar, Agrim Gupta, Xiuye Gu, Alexander~G. Hauptmann, Boqing Gong, Ming-Hsuan Yang, Irfan Essa, David~A. Ross, and Lu~Jiang.
\newblock Language model beats diffusion -- tokenizer is key to visual generation.
\newblock In \emph{ICLR}, 2024.

\bibitem[Zhang et~al.(2025)Zhang, Chen, Xu, Wang, Yang, Meng, Guo, Zhao, and Zhang]{zhang2025panodit}
Muyang Zhang, Yuzhi Chen, Rongtao Xu, Changwei Wang, JinMing Yang, Weiliang Meng, Jianwei Guo, Huihuang Zhao, and Xiaopeng Zhang.
\newblock Pano{D}it: Panoramic videos generation with diffusion transformer.
\newblock In \emph{Proc. of the AAAI Conference on Artificial Intelligence}, 2025.

\bibitem[Zhou et~al.(2025)Zhou, Yu, Guan, Cheng, Tian, and Yuan]{zhou2025holotime}
Haiyang Zhou, Wangbo Yu, Jiawen Guan, Xinhua Cheng, Yonghong Tian, and Li~Yuan.
\newblock Holotime: Taming video diffusion models for panoramic 4d scene generation, 2025.
\newblock URL \url{https://arxiv.org/abs/2504.21650}.

\end{thebibliography}

\end{document}